\documentclass[preprint, 12pt, a4paper]{elsarticle}
\usepackage{amssymb}
\usepackage{lineno}
\usepackage{xcolor}

\usepackage[frozencache=true,cachedir=.]{minted} 
\setmintedinline{breaklines}
\usepackage{xurl}
\usepackage{hyperref}

\usepackage{float}
\restylefloat{table}
\begin{document}

\begin{frontmatter}

\title{Scilab-RL: A software framework for efficient
reinforcement learning and cognitive modeling research}

\author[]{Jan Dohmen}
\ead{jan.dohmen@tuhh.de}
\author[]{Frank Röder}
\ead{frank.roeder@tuhh.de}
\author[]{Manfred Eppe}
\ead{manfred.eppe@tuhh.de}
\address{Institute for Data Science Foundations \\ Hamburg University of Technology\\ Hamburg, Germany}

\begin{abstract}
One problem with researching cognitive modeling and reinforcement learning (RL) is that researchers spend too much time on setting up an appropriate computational framework for their experiments. Many open source implementations of current RL algorithms exist, but there is a lack of a modular suite of tools combining different robotic simulators and platforms, data visualization, hyperparameter optimization, and baseline experiments.
To address this problem, we present Scilab-RL, a software framework for efficient research in cognitive modeling and reinforcement learning for robotic agents. 
The framework focuses on goal-conditioned reinforcement learning using Stable Baselines 3 and the OpenAI gym interface. 
It enables native possibilities for experiment visualizations and hyperparameter optimization.
We describe how these features enable researchers to conduct experiments with minimal time effort, thus maximizing research output.

\end{abstract}


\end{frontmatter}

\section*{Code metadata}
\label{metadata}

\begin{table}[H]
\begin{tabular}{|l|p{5.8cm}|p{5.8cm}|}
\hline
\textbf{Nr.} & \textbf{Code metadata description} & \textbf{Please fill in this column} \\
\hline
C1 & Current code version & v0.2.0\\
\hline
C2 & Permanent link to code/repository used for this code version & \url{https://github.com/Scilab-RL/Scilab-RL} \\
\hline
C3 & Legal Code License  & MIT \\
\hline
C4 & Code versioning system used & Git \\
\hline
C5 & Software code languages, tools, and services used & Python, Shell \\
\hline
C6 & Compilation requirements, operating environments \& dependencies & See requirements.txt, mujoco-py, PyTorch\\
\hline
C7 & If available, link to developer documentation/manual & \url{https://scilab-rl.github.io/Scilab-RL/} \\
\hline
C8 & Support email for questions & \url{https://github.com/Scilab-RL/Scilab-RL/issues}\\
\hline
\end{tabular}
\caption{Code metadata}
\label{code:metadata} 
\end{table}

\section{Motivation and significance}
\label{motivation}

Breakthroughs such as AlphaGo \cite{silver_mastering_2016}, solving Atari games \cite{mnih_playing_2013}, and most recently, the discovery of new approaches to faster matrix multiplication algorithms \cite{fawzi_discovering_2022} are contributing to the growing popularity of reinforcement learning (RL).

The development of new RL methods involves an iterative process of repeated design, testing, and analysis. To accelerate this process, researchers need a framework that includes a variety of robust RL algorithms and environments for benchmarking, and also possibilities for visualization, hyperparameter optimization and testing. Developing an integrated, robust workflow where all these tools synergize with each other is time-consuming and distracts from the actual research question in mind. Interdisciplinary researchers that are non-native programmers suffer most from this problem, even though the field of artificial intelligence can benefit tremendously from theories developed in for example psychology or cognitive science. 

To address this problem, there exist computational frameworks for the rapid prototyping of new RL approaches and methods. 
As examples, consider SpinningUp \cite{achiam_spinning_2018}, CleanRL \cite{huang_cleanrl_2022}, Garage \cite{the_garage_contributors_garage_2019}, RLlib \cite{liang_rllib_2018}, Tianshou \cite{weng_tianshou_2022} or RL Baselines3 Zoo \cite{raffin_rl_2020}.
Some of them can be considered a very valuable entry point into RL, as their codebase is lightweight, concise and understandable \cite{huang_cleanrl_2022,weng_tianshou_2022} or their documentation itself can be considered as an educational resource to learn about RL \cite{achiam_spinning_2018}. Their strengths lie, for example, in supporting a wide range of algorithms \cite{huang_cleanrl_2022,the_garage_contributors_garage_2019,liang_rllib_2018,weng_tianshou_2022,raffin_rl_2020}, in performance \cite{liang_rllib_2018}, or in reproducibility \cite{the_garage_contributors_garage_2019}. However, some are missing a direct way of integrating supportive tools e.g. for hyperparameter optimization \cite{the_garage_contributors_garage_2019,weng_tianshou_2022}. While options to support cloud-based experiment tracking and metric logging are quite common among this kind of frameworks \cite{achiam_spinning_2018,huang_cleanrl_2022,the_garage_contributors_garage_2019,liang_rllib_2018,weng_tianshou_2022,raffin_rl_2020}, an additional aspect that we have found to be very valuable for RL development, but that to the best of our knowledge we could not find anywhere else, is the ability to observe metrics such as Q-values or rewards in parallel with the rendering of the environment.

In that regard, we propose Scilab-RL, a robust framework for efficient experimenting in RL. The framework is especially addressed to new researchers in the fields of RL and cognitive modeling and aimed at providing an easier access to computational modeling. In the context of robotic tasks we focus on methods that avoid reward-shaping, by using goal-conditioned reinforcement learning \cite{schaul_universal_2015} and hindsight experience replay \cite{andrychowicz_hindsight_2017} instead. A core feature is the aforementioned visualization of metrics during runtime, as illustrated in Figure \ref{fig:fetchpush_experiment}.
Overall, our framework involves the following features:
\begin{enumerate}[a)]
    \item A standardized sense-act interface based on OpenAI gym \cite{brockman_openai_2016}, using the robotic simulators MuJoCo \cite{todorov_mujoco_2012} and CoppeliaSim \cite{rohmer_v-rep_2013}.
    \item A hyperparameter optimization engine based on Optuna \cite{akiba_optuna_2019}.
    \item Implementations of the currently most important RL algorithms based on Stable Baselines 3
    \cite{raffin_stable-baselines3_2021}.
    \item Multiple data visualization methods, including experiment monitoring using MLFlow \cite{zaharia_accelerating_2018} or Weights \& Biases \cite{biewald_experiment_2020} and online visualization of metrics, like Q-values or intrinsic rewards.
    \item Testing scripts for continuous integration and delivery.
\end{enumerate}
To obtain Scilab-RL, visit our GitHub repository (see Table \ref{code:metadata}). The tool is designed to be executed from the command line, with configurations specified in dedicated setting files. After running an experiment, the results can be analyzed using the integrated visualization capability or by linking to external visualization tools.

\pagebreak
\section{Software description}
\label{description}

\subsection{Software architecture}
\label{achitecture}

The core of Scilab-RL, which is written in Python, leverages YAML files to configure the algorithms and shell scripts to define the overall setup and testing procedures.
It integrates RL algorithms from Stable Baselines 3 \cite{raffin_stable-baselines3_2021}, allowing straightforward pairing with the OpenAI Gym interface \cite{brockman_openai_2016} for environment interactions. Furthermore, robotic simulators such as MuJoCo \cite{todorov_mujoco_2012} and CoppeliaSim \cite{rohmer_v-rep_2013} define the framework's wide variety of mostly goal-conditioned benchmark tasks. 

\begin{figure}[ht]
    \centering
    \includegraphics[width = \textwidth]{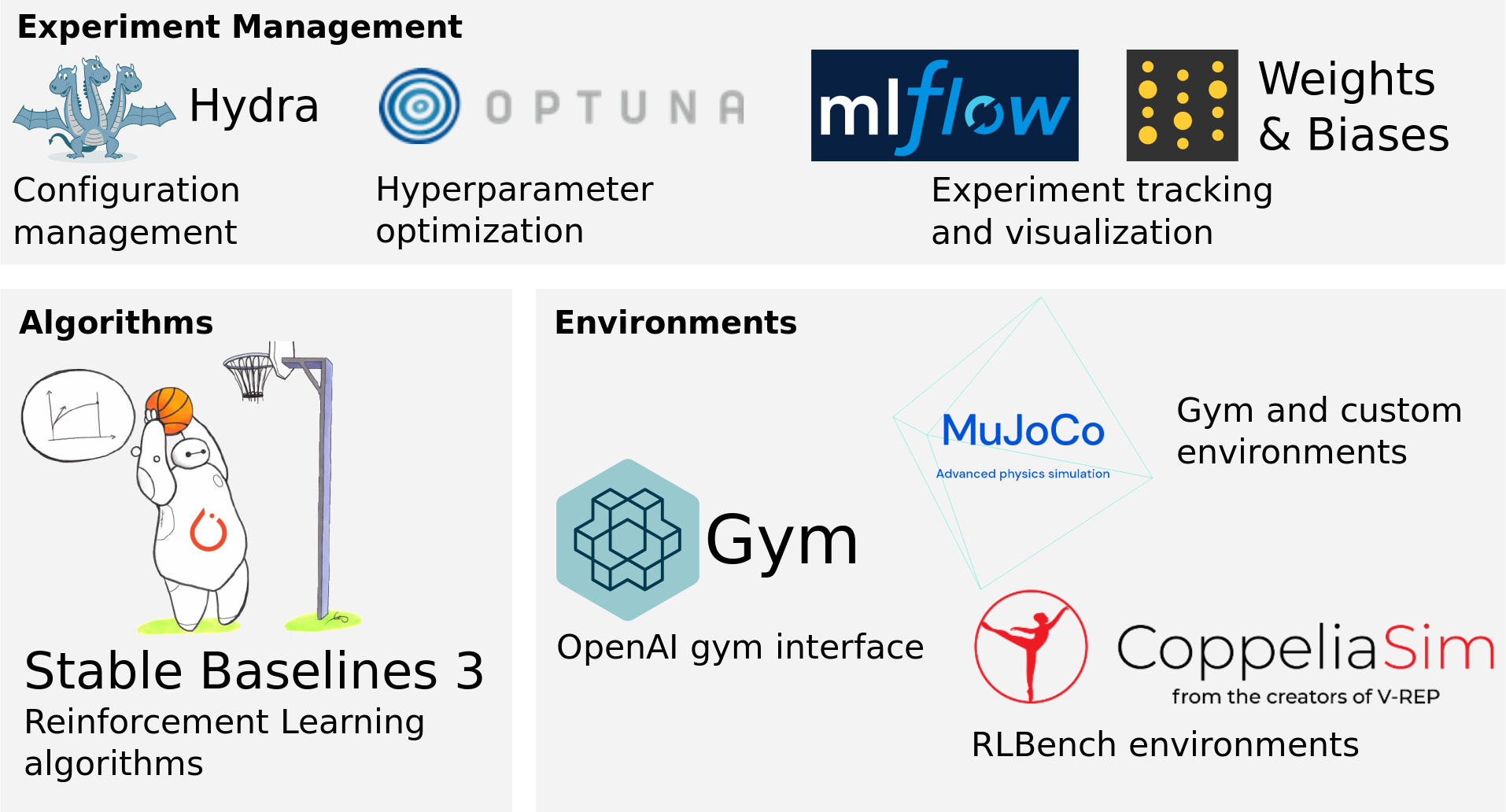}
    \caption{An overview of the tools used in Scilab-RL.}
    \label{fig:overview}
\end{figure}

The framework incorporates specialized tools and customized functionalities to facilitate intuitive experiment management and evaluation. These include configuration management, hyperparameter optimization, experimental tracking, and visualization.
Figure \ref{fig:overview} and Section \ref{funcionalities} provide an overview of the Scilab-RL components.

\subsection{Software functionalities}
\label{funcionalities}
The following sections present the core functionalities. However, we recommend users to examine our dedicated Wiki \footnote{\url{https://scilab-rl.github.io/Scilab-RL/}} for additional tutorials, examples, and a detailed documentation.

\subsubsection{RL algorithm development}
Research on RL and cognitive modeling involves testing hypotheses by implementing new or modified RL algorithms. With our Stable Baselines 3 integration, this becomes straightforward using the Scilab-RL framework. The only requirement is that the developed algorithms follow the sense-act interface based on OpenAI gym. The pivotal benefits that we offer, are the integrated data visualization, hyperparameter optimization, online rendering and tests.

\subsubsection{Environment development}
Our framework involves a variety of existing RL benchmark environments, particularly environments with sparse rewards for goal-conditioned RL to avoid reward-shaping. 
However, it is often important for a research question to develop new environments. Using Scilab-RL speeds up this task because developers do not need to spend time on implementing RL algorithms or environments.

\subsubsection{Visualization}
Developing and debugging new algorithms or environments can be quite challenging and time-consuming. Here, the visualization capability can be of great importance, e.g. for discovering functional flaws early on.
We feature two kinds of data visualization. The first kind of visualization is common in similar RL-focused frameworks. It involves the cloud-based logging and tracking of metrics using MLFlow \cite{zaharia_accelerating_2018} and Weights \& Biases \cite{biewald_experiment_2020}. This enables researchers to review ongoing experiments. For example, one can assess the learning performance during hyperparameter optimization, even when experiments are running on different machines. 

The second kind of visualization is usually not available in similar frameworks, but it can be of immediate help: the online visualization of user-defined metrics in sync with the rendering provided by the respective environment. Developers can define and visualize any number of online metrics at runtime. In Section \ref{example}, we use the online metric visualization to make sense of the agent's behavior and thus verify a proposed hypothesis.

\subsubsection{Hyperparameter optimization}
Reinforcement learning typically involves many hyperparameters that are hard to manage and optimize. To address this issue, we use Hydra \cite{yadan_hydra_2019}, a robust and versatile hyperparameter management system that connects seamlessly with hyperparameter optimization frameworks such as Optuna \citep{akiba_optuna_2019}. Default parameter configurations are stored in corresponding YAML files and are tied to an algorithm or an algorithm-environment combination.
This makes it especially useful if different environment require different parameters.
Furthermore, we allow alternating or adding individual parameter values.

\subsubsection{Tests}
Testing is important for continuous integration and delivery pipelines. For example, when underlying packages receive an update, or a common utility function changes in the code, Scilab-RL offers performance and smoke tests to verify these. Smoke tests check the pure functionality, whether everything within the framework still runs. The performance tests verify that our RL algorithm's performance are not affected negatively by changes to the code.

\section{Illustrative example}
\label{example}

A major challenge in RL is exploration \cite{pathak_curiosity-driven_2017,burda_exploration_2018}. Using Scilab-RL, we provide an example workflow on how to modify an existing state-of-the-art RL algorithm (Soft-Actor Critic, SAC \cite{haarnoja_soft_2018}) to enhance its capability of exploration. A key feature of SAC is that it incorporates the entropy, or roughly speaking the uncertainty of the agent's policy (actor) as regularization for the optimization problem. This directly motivates our hypothesis that we want to test in the following use case. Not just the uncertainty of the policy but also the uncertainty of the Q function could be used to improve the learning behavior of an agent. A high variance of an ensemble of critics (Q-values) could imply that the agent is uncertain about its current situation and is exploring new state-action pairings in the environment. We assume that encouraging higher variances of critics improves exploration. A straightforward way to test this hypothesis is to use the variance of $N$ critics as an additional intrinsic reward $r_i$ (\autoref{eq:intrinsic}), where the variance $\mathrm{Var}[\cdot]$ is defined as the average of the squared deviations from the mean.

\begin{equation}
    \label{eq:intrinsic}
    r_i = \mathrm{Var}[Q_{\Phi1}(s, a), Q_{\Phi2}(s, a), \dots Q_{\Phi N}(s, a)]
\end{equation}

We perform the following steps: 
First, we clone the Scilab-RL repository. Within the \mintinline{text}{src/custom_algorithms} folder, we create a copy of the SAC algorithm implementation of Stable Baselines 3 and rename every \mintinline{text}{sac} to a custom name, e.g. \mintinline{text}{sac_var}.
In addition, we create a respective algorithm configuration file \mintinline{text}{conf/algorithm/sac_var.yaml} for the hydra hyperparameter management system. 

Now, we would be able to use the new (copied) algorithm from the command line via:
\mint{text}{python src/main.py algorithm=sac_var env=FetchPush-v1}  
Next, we modify the algorithm to add the intrinsic rewards. 
Within the created \mintinline{text}{sac_var.py} file in the \mintinline{text}{SAC_VAR} class, we change the \mintinline{text}{train()} method so that the variance of the critics is computed. Here we add it as intrinsic reward $r_i$ (\autoref{eq:intrinsic}) to the extrinsic reward $r_e$, using a weight $\eta$ according to \autoref{eq:reward}. This modified reward is then used to compute the target Q-values with the common Bellman updates. An example algorithm modification is provided as a code excerpt in \ref{code:sac}.

\begin{equation}
    \label{eq:reward}
    r = (1 - \eta) r_e + \eta r_i
\end{equation}

\begin{figure}[ht]
    \centering
    \includegraphics[width=\textwidth]{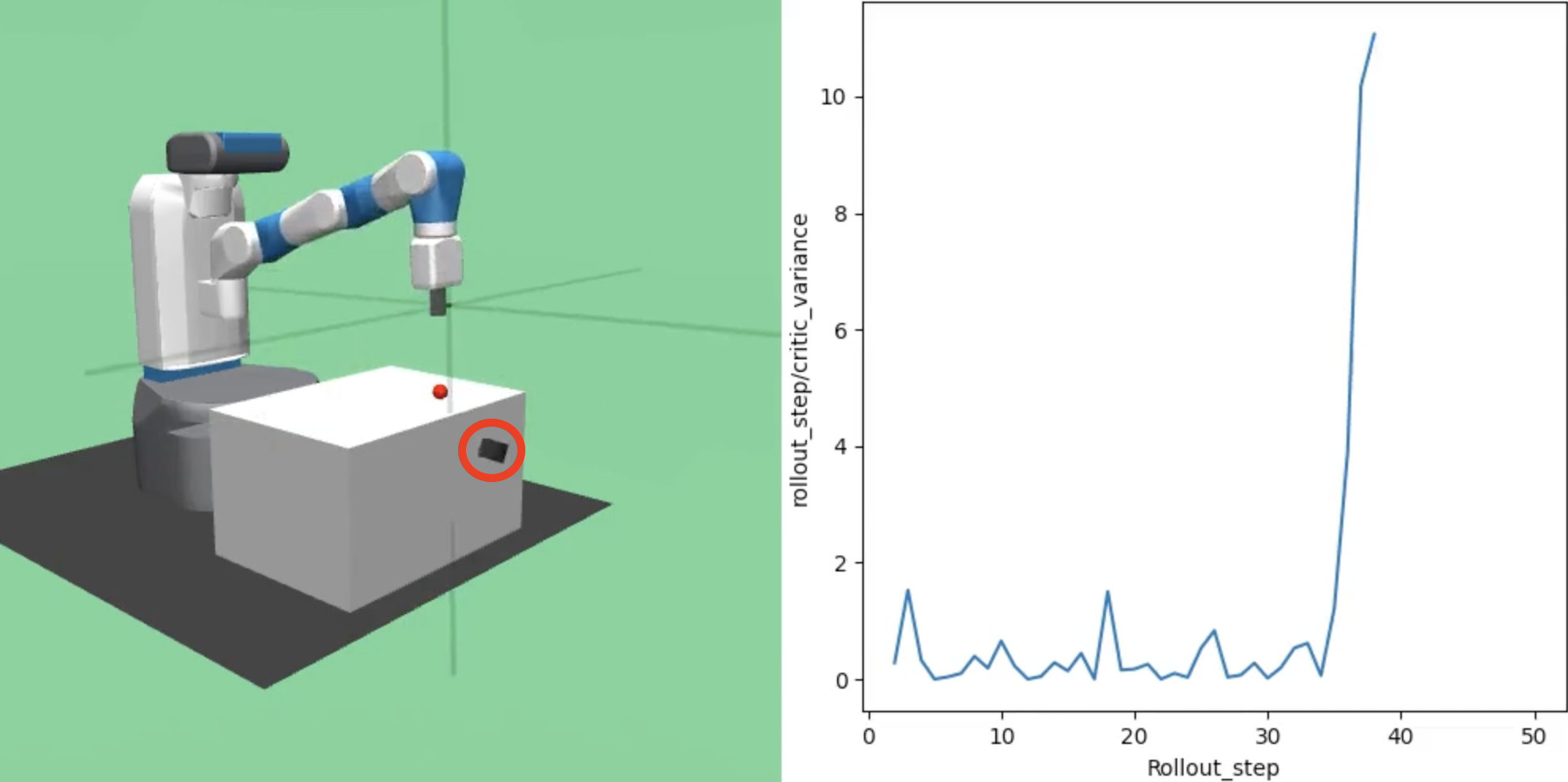}
    \caption{Online rendering capabilities. Example using the MuJoCo FetchPush environment.}
    \label{fig:fetchpush_experiment}
\end{figure}

After implementing the novel intrinsic reward, it is desirable to verify that the implementation operates as intended. We can do this verification with the online metric visualization, as depicted in \autoref{fig:fetchpush_experiment}. It shows the rendered visualization of the environment and a corresponding user-defined metric value, in this case the variance of the critics, side by side. In this depicted scenario, the robotic arm accidentally pushes the block (highlighted red) over the edge of the platform, causing it to fall to the ground. As these kinds of situations are rather uncommon, the critics can not reliably predict, hence we would expect their variance to be high. This is confirmed by the large increase in variance (right of Figure \ref{fig:fetchpush_experiment}), and indicates that our implementation indeed works as intended. In this regard, online visualization helps to examine the hypothesis made at the beginning of this example that critics' uncertainty may be an indicator of unusual situations and that rewarding uncertainty may encourage exploration.

Having verified that our implementation works as intended, we perform hyperparameter optimization. Here, we optimize the parameter \newline \mintinline{text}{weight_critic_var} ($\eta$ in \autoref{eq:reward}) for which we want to choose values categorically from [0.0, 0.25, 0.5, 0.75, 1.0]. An example code excerpt for the hyperparameter optimization setting is provided in \ref{code:conf}.
The optimization maximizes the agent’s \mintinline{text}{success_rate}. The \mintinline{text}{success_rate} describes the proportion of episodes in which the agent terminates in the desired goal state, i.e. was successful \cite{andrychowicz_hindsight_2017}.

\begin{figure}[ht]
    \centering
    \includegraphics[width=\textwidth]{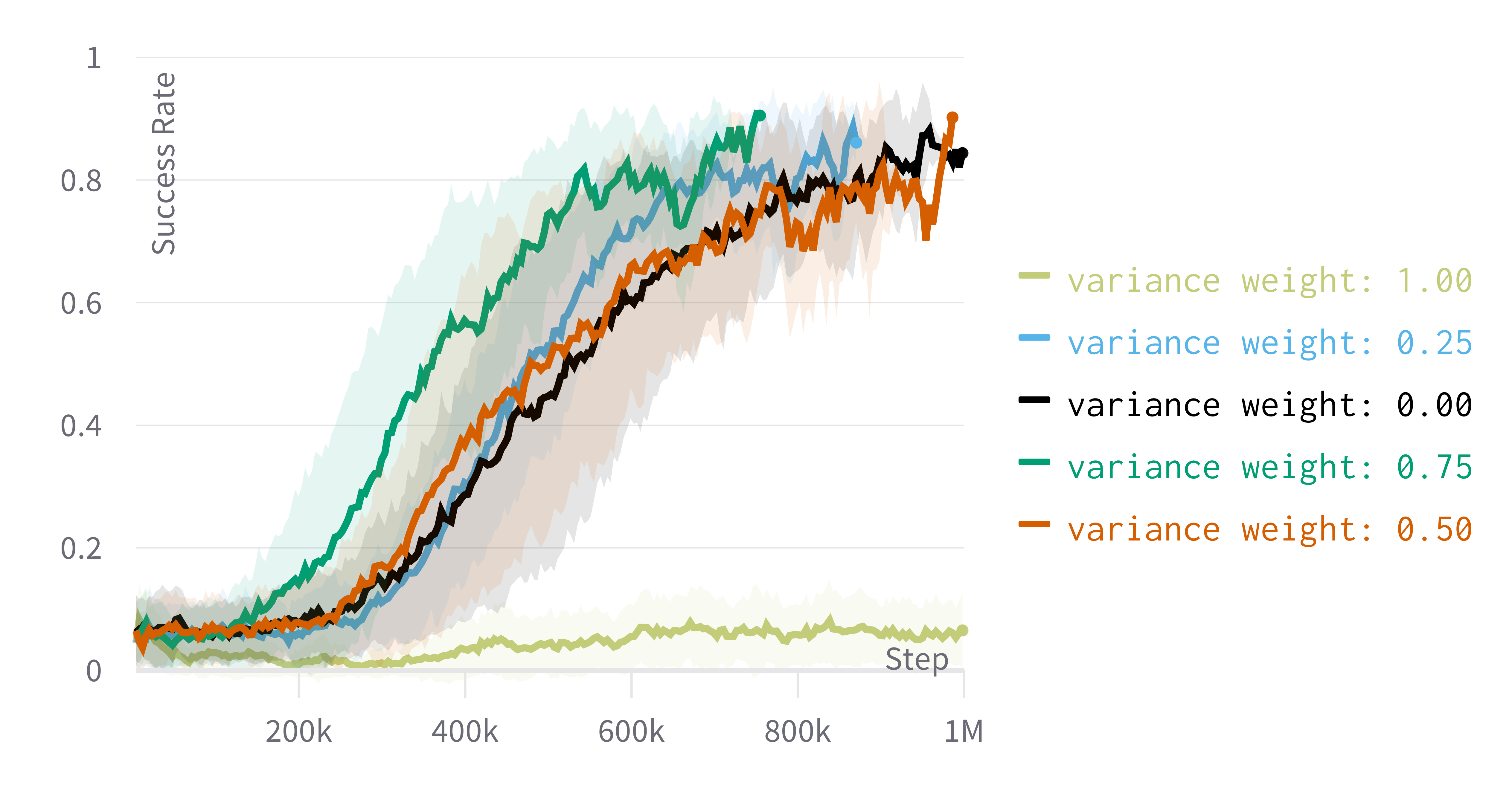}
    \caption{Hyperparameter optimization for the modified SAC algorithm adding critic variance to the reward. Experiment for the FetchPush environment.}
    \label{fig:var_experiment}
\end{figure}

Figure \ref{fig:var_experiment} shows how one can track information on the training progress and evaluation metrics live via the linkage to Weights \& Biases. For this example, \mintinline{text}{weight_critic_var = 0.75} leads to the best results. The graph directly verifies our hypothesis that adding the variance of the critics as an intrinsic reward indeed improves the learning performance. 
The entire process from setting up the experiment to obtaining results did not require much effort overall using Scilab-RL.

\section{Impact}
\label{impact}
In cognitive modeling and reinforcement learning, setting up an appropriate computational framework for experiments can be a tedious and time-consuming task. Research in RL is quite a matter of trial and error, and the frequency of trying new ideas and testing hypotheses should ideally be high. In addition, it is important to create easier access for non-ML experts and reduce their programming overhead, as the field of AI is quite interdisciplinary and can benefit from a variety of fields such as psychology and cognitive science.

Scilab-RL helps researchers to get started with goal-conditioned RL for robotic agents, and lets them focus on efficient experimenting. The framework is tailored towards the rapid prototyping, development and evaluation of new RL algorithms, methods and environments. This is achieved by bringing together various state-of-the-art RL algorithms, environments, the possibility for hyperparameter optimization and a built-in data visualization for fast and efficient debugging and evaluation. The setup effort is reduced to a minimum so that first agents can be trained right away. The framework is command line based, while it provides a systematic management of configurations for different experiments. Approaches are directly implemented in the cloned repository. Scilab-RL is particularly intended for researchers new to the field, but experts can also benefit as the underlying algorithms and tools are state-of-the-art and constantly updated.

\section{Conclusions}
\label{conclusions}
We present Scilab-RL, a reinforcement learning framework that accelerates simple and user-friendly experimenting with a minimum of coding effort and setup required. In this paper, we introduced the key features and architecture of the framework and demonstrated its potential usage by means of a typical illustrative example arising from a relevant research question. 

Scilab-RL is open for contributions and encourages users to integrate established approaches and algorithms or new challenging environments into the framework and thus make them usable for others.

\section{Conflict of interest}
No conflict of interest exists:
We wish to confirm that there are no known conflicts of interest associated with this publication and there has been no significant financial support for this work that could have influenced its outcome.

\section*{Acknowledgements}
\label{acknowledgements}
The authors gratefully acknowledge funding by the German Research Foundation DFG through the MoReSpace (402776968) and LeCAREbot \linebreak (433323019) projects. Additionally, we would like to thank Thilo Fryen and Matin Urdu for their contributions to the codebase.
\appendix

\section{Code excerpts for the illustrative example}
\subsection{Modified SAC algorithm}
\label{code:sac}

\begin{minted}{python}
class SAC_VAR(OffPolicyAlgorithm):
    def __init__(
        self,
        ... # left out for clarity
        weight_critic_var: float,
        
    ):
        super(SAC_VAR, self).__init__(
            ... # left out for clarity
        )
        self.weight_critic_var = weight_critic_var
        
    def train(
        self, 
        gradient_steps: int, 
        batch_size: int = 64
    ) -> None:
        # compute the variance of Q values
        q_values = th.cat(
            self.critic_target(
                replay_data.next_observations, next_actions
            ), 
            dim=1
        )
        critic_variance = q_values.var(dim=1)
        # min-max scale variance between 0 and 1
        var_max = th.max(critic_variance)
        var_min = th.min(critic_variance)
        critic_variance = (
            (critic_variance - var_min) / (var_max - var_min)
        )
        critic_variance = critic_variance.unsqueeze(1)
        critic_variances.append(critic_variance.tolist())
        # extrinsic + intrinsic reward (weighted) 
        reward_mod = (
            (1 - self.weight_critic_var) * replay_data.rewards \
            + self.weight_critic_var * critic_variance
        )
        # compute target Q-values
        target_q_values = (
            reward_mod + (1 - replay_data.dones) \
            * self.gamma * next_q_values
        )
        ... # left out for clarity
\end{minted}

\subsection{Hydra sweeper configuration}
\label{code:conf}
\begin{minted}{yaml}
env: 'FetchPush-v1'
hydra:
  sweeper:
    study_name: sac_var_FetchPush
    max_trials: 64
    n_jobs: 16
    direction: maximize
    min_trials_per_param: 3
    max_trials_per_param: 12
    search_space:
      ++algorithm.weight_critic_var:
        type: categorical
        choices:
          - 0.0
          - 0.25
          - 0.5
          - 0.75
          - 1.0
\end{minted}

\bibliographystyle{elsarticle-num}
\bibliography{ScilabRL}

\end{document}